\documentclass[mnsc, nonblindrev]{informs3}

\OneAndAHalfSpacedXI

\usepackage{natbib}
 \bibpunct[, ]{(}{)}{,}{a}{}{,}%
 \def\bibfont{\small}%
\usepackage{hyperref}
\hypersetup{
	colorlinks,
	linkcolor=red,
  citecolor=blue,
  urlcolor=black
}
\usepackage{multicol}
\usepackage{multirow}
\usepackage{booktabs}
\usepackage{subfigure}
\usepackage{lmodern}
\newcolumntype{L}[1]{>{\raggedright\let\newline\\\arraybackslash\hspace{0pt}}m{#1}}
\newcolumntype{C}[1]{>{\centering\let\newline\\\arraybackslash\hspace{0pt}}m{#1}}
\newcolumntype{R}[1]{>{\raggedleft\let\newline\\\arraybackslash\hspace{0pt}}m{#1}}
\TheoremsNumberedThrough     
\ECRepeatTheorems

\EquationsNumberedThrough    

\begin{document}


\RUNAUTHOR{Du et al.}

\RUNTITLE{Exploring a New Competency Modeling Process with Large Language Models}

\TITLE{Exploring a New Competency Modeling Process with Large Language Models}

\ARTICLEAUTHORS{%
\AUTHOR{Silin Du, Manqing Xin}
\AFF{School of Economics and Management, Tsinghua University, Beijing, China\\
\EMAIL{dsl21@mails.tsinghua.edu.cn}, \EMAIL{xmq21@mails.tsinghua.edu.cn}} 
\AUTHOR{Raymond Jia Wang}
\AFF{Bill-JC Technology, Wuhan, China \\ \EMAIL{wangjia@bill-jc.com}}
{\centering
Last Update: \today
}
} 

\ABSTRACT{%

Competency modeling is widely used in human resource management to select, develop, and evaluate talent. However, traditional expert-driven approaches rely heavily on manual analysis of large volumes of interview transcripts, making them costly and prone to randomness, ambiguity, and limited reproducibility.  This study proposes a new competency modeling process built on large language models (LLMs). Instead of merely automating isolated steps, we reconstruct the workflow by decomposing expert practices into structured computational components. Specifically, we leverage LLMs to extract behavioral and psychological descriptions from raw textual data and map them to predefined competency libraries through embedding-based similarity. We further introduce a learnable parameter that adaptively integrates different information sources, enabling the model to determine the relative importance of behavioral and psychological signals. To address the long-standing challenge of validation, we develop an offline evaluation procedure that allows systematic model selection without requiring additional large-scale data collection. Empirical results from a real-world implementation in a software outsourcing company demonstrate strong predictive validity, cross-library consistency, and structural robustness. Overall, our framework transforms competency modeling from a largely qualitative and expert-dependent practice into a transparent, data-driven, and evaluable analytical process.
}%



\maketitle

%


\section{Introduction}

How to select, retain, and cultivate talent is fundamental to maintaining an organization's competitiveness. To address this challenge, many organizations utilize the notion of competencies as an integrated framework to understand what enables effective job performance. Competencies are typically defined as the underlying knowledge, skills, abilities, and other characteristics that distinguish superior from average performers~\citep{boyatzis1991competent,campion2011doing}. Building on this idea, competency modeling, which refers to the systematic process of identifying, describing, and organizing the competencies required for successful performance in a specific job, role, or organizational context, has become an effective and popular tool~\citep{Shippmann2000}. In today’s organizations, competency models form the foundation for various human resource management (HRM) activities and are an indispensable tool for talent management \citep{cappelli2014talent}. However, along with the widespread adoption of competency models in the industry, many criticisms and challenges have emerged, particularly concerning the lack of rigor in the modeling process and the associated costs \citep{Shippmann2000,Stevens2013}.

Traditional competency modeling processes often suffer from substantial randomness and ambiguity in identifying competencies, resulting in low reproducibility \citep{Shippmann2000}. Much of this problem comes from the traditional expert-driven workflow. A group of experts needs to interview participants, read interview transcripts, extract information, and summarize the content into competency statements. Such qualitative and expert-dependent steps place high demands on human information processing and make the outcomes highly sensitive to the individual experts involved. Different groups of experts may produce various results. In addition, competency modeling is highly resource-intensive~\citep{campion2011doing}. Conducting interviews, analyzing large amounts of transcripts, and reaching expert consensus require substantial time and specialized expertise. Many organizations cannot easily commit such resources. In practice, attempts to improve model quality, such as increasing interview coverage, only intensify these demands and further inflate the overall cost. 

Another major limitation is that it is extremely difficult to validate a competency model within the traditional process. Validating a competency model requires collecting additional competency data from other employees to verify whether the identified competencies truly distinguish high and average performers. Organizations may need to conduct more interviews, analyze additional transcripts, distribute questionnaires, or run assessment center activities. However, all of these methods require substantial human effort, time, and coordination. For this reason, validation is frequently omitted or carried out only superficially, leaving many competency models as one-off, unverified products with uncertain reliability.

The expert-driven process also faces challenges in adaptability. Once a competency model is established, it is hard to adjust its structure (e.g., the number of key competencies), limiting the model’s ability to evolve with changing organizational needs. Although prior research has suggested using competency libraries\footnote{Competency libraries refer to a list of competencies available for selection during the development of a competency model. These competencies often represent experiences derived from other companies or other projects within the same company. Utilizing a competency library can simplify and expedite the development of a competency model. Table \ref{tab1} shows the structure of a competency library and some examples.} or establishing advisory panels, the current method for competency modeling continues to face significant drawbacks.

Given these limitations, there is a strong need for a more rigorous, objective, and efficient way to conduct competency modeling. Many issues above stem directly from the heavy reliance on experts to read, interpret, and summarize interview transcripts. These limitations highlight the need for a more scalable and consistent analytical mechanism. The rapid development of large language models (LLMs) offers a promising technological pathway. Modern LLMs, such as GPT-4 \citep{achiam2023gpt}, demonstrate exceptional abilities in understanding and generating human language \citep{Zhao2023}. Empirical studies have shown that LLMs can perform complex and nuanced tasks, such as identifying non-answers in earnings conference calls \citep{de2025chatgpt}. These capabilities suggest the potential of LLMs for automating complex qualitative analyses in competency modeling.

Recent advances have further expanded the context windows of LLMs to more than 100k tokens \citep{Peng2023}, enabling LLMs to process entire interview transcripts at once. This addresses one of the major bottlenecks of traditional approaches: the difficulty of systematically analyzing large volumes of textual data. Furthermore, LLMs can generate outputs rapidly and at relatively low marginal cost through API services. This reduces the dependence on scarce expert resources and lowers the barrier for small organizations to carry out competency modeling.

Importantly, LLMs make validation and iterative refinement far more feasible. LLMs can analyze interview transcripts quickly and consistently, which allows organizations to set aside a portion of interview data as a test set. After a competency model identifies key competencies that should separate high and low performers, the LLM can score the test-set interviews on these competencies. These scores can then be compared directly to the actual performance levels of the individuals in the test set. This creates a fast, low-cost way to validate whether a competency model truly works, without requiring experts to repeat large amounts of manual analysis. LLMs thus enable offline, systematic evaluation of different modeling choices and support rapid refinement of the overall workflow.

Despite these promising opportunities, an open question remains: how should LLMs be integrated into the competency modeling workflow, and at which stages can they provide the greatest value while preserving interpretability and evaluability? Addressing this question requires a careful redesign of the competency modeling process to harmonize LLM capabilities with established HR practices. 

To this end, we design a new \underline{co}mpetency modeling process with \underline{l}arge \underline{l}anguage \underline{m}odels, namely \textbf{CoLLM}. The contributions of this paper are threefold.
\begin{enumerate}
    \item To the best of our knowledge, this is the first work to use LLMs to redesign the entire competency modeling workflow. We make this core HRM tool more rigorous and efficient, and to enable more organizations to apply it at a low cost.
    \item We propose a new LLM-based competency modeling process. First, we use in-context learning to prompt LLMs to summarize behavioral and psychological descriptions from the raw transcripts. Then we map the extracted data into the competency library with the embedding models. Additionally, we design a learnable weight to adaptively integrate the behavioral and psychological data. Finally, we devise an offline evaluation process for model selection. 
    \item To verify the effectiveness of the proposed method, we collect real-world data from a software outsourcing company and conduct extensive experiments and manual comparison.
\end{enumerate}

\section{Related Work}
\subsection{Competency Modeling}
The competency model refers to a collection of knowledge, skills, abilities, and other characteristics necessary for effective performance in a relevant job \citep{campion2011doing}. Competency modeling is an attribute-based form of work analysis that aims to identify the competencies required for specific roles~\citep{Stevens2013}. Competency modeling can be traced back to the 1970s and is often attributed to the work of \citet{McClelland1973}. He argued that traditional intelligence and personality tests were inadequate in effectively capturing and predicting workplace performance and proposed shifting the focus to competencies. His approach to comparing high-performing and average-performing individuals deeply influenced the practice of competency modeling. The techniques used in the competency modeling process significantly overlap with those used in job analysis ~\citep{Shippmann2000}. The key differences between competency modeling and job analysis lie in their strategic relevance and their ability to distinguish between high performance and average performance \citep{campion2011doing}.


How is a competency model developed? \citet{briscoe1999grooming}, through interviews with 31 leading North American companies, identified three primary foundations for building competency models: research-based, strategy-based, and values-based. \citet{campion2011doing} systematically outlined the best practices for competency modeling from a full-cycle perspective, encompassing three topics: Analyzing Competency Information (Identifying Competencies), Organizing and Presenting Competency Information, and Using Competency Information. Their approach reflects the necessity of integrating research-based, strategy-based, and values-based approaches in developing competency models. In practice, a highly valuable tool is the competency library. Over years of practical experience, the industry has developed several competency libraries that use standardized language to describe various dimensions of competencies, including skills, knowledge, abilities, etc. Competency libraries are typically hierarchical, offering terminology at varying levels of granularity to describe competencies (as illustrated in Table \ref{tab1}). Organizations are strongly advised to select and utilize a competency library prior to modeling, as this helps establish a comparable conceptual hierarchy and enhance efficiency~\citep{campion2011doing}.

\subsection{Large Language Models}
Recently, both academia and industry have made significant progress in LLMs ~\citep{achiam2023gpt, chang2024survey}. LLMs, distinct from pre-trained language models, are characterized by an enormous number of parameters (usually over 10 billion). As the model size and data size scale up, LLMs have demonstrated emergent abilities \citep{Wei2022}. The technical evolution of LLMs has significantly reshaped AI research and expanded their applicability beyond traditional NLP tasks to complex reasoning, strategic interaction, and structured decision-making settings 
\citep{Zhao2023,du2024listwise,du2024helmsman,che2024coldstart}. 

Low reproducibility and inefficiency in competency modeling stem from the requirement for experts to meticulously analyze a large volume of interview transcripts. Fortunately, LLMs exhibit great abilities to generate summarization, and the quality of LLM-generated summaries is judged on par with those written by humans \citep{Zhang2024}. Meanwhile, the text embedding technique has also made significant advancements thanks to LLMs. \citet{Neelakantan2022} initialize word embeddings with GPT-3 \citep{Brown2020} and use contrastive training to yield high-quality word representations. Besides, the context limit of LLMs has been extended to over 100k~\citep{Peng2023}, offering basic conditions to handle extremely long transcripts. Thus, in this paper, we explore a new process for competency modeling equipped with LLMs, which first utilizes LLMs to summarize related descriptions from the raw texts and then uses the text similarity based on the embedding technique to construct competency models.

\section{Proposed Method}
\subsection{Traditional Competency Modeling Process}


To begin with, we present a formal definition of competencies as follows.
\begin{definition}[Competency]
    A competency is an underlying characteristic of an individual that enables effective or superior performance in a job or role, such as knowledge, skills, abilities, motives, or behavioral tendencies~\citep{McClelland1973}.
\end{definition}
Although competencies incorporate elements like skills and abilities, the concept differs in important ways. Skills refer to learned capabilities to perform specific tasks, and abilities represent innate or general capacities. Competencies, in contrast, integrate these elements with behavioral patterns and personal attributes, emphasizing how individuals actually perform in real work situations rather than what they can do in theory. As such, competencies reflect a more holistic and context-dependent understanding of performance. Based on this, we define the goal of competency modeling as follows.
\begin{definition}[Competency Modeling]
    The goal of competency modeling is to identify, describe, and organize the competencies required for effective performance in a specific job, job family, or organizational domain \citep{Shippmann2000}.
\end{definition}

A competency model typically highlights the characteristics that distinguish high performers from average performers and provides a structured framework for aligning talent selection, development, performance appraisal, and succession planning. Currently, the tools used to develop competency models primarily include job analysis interviews, behavioral event interviews (BEI), questionnaires, and the integration of existing competency libraries. Each method has its advantages and disadvantages, and the choice of specific application processes should consider the organizational context. Nevertheless, both theory and practice acknowledge the effectiveness and importance of the behavioral event interview method. In the social sciences, where human subjects are the focus of study, individuals often alter their behavior upon learning they are being observed, attempting to meet perceived researcher expectations and consequently exhibiting atypical or unintended responses that compromise research outcomes. BEI method effectively mitigates this issue by ensuring the authenticity and richness of collected data. BEI is a structured interview technique designed to elicit competencies by exploring specific, real-world situations and detailed behavioral accounts. The BEI process requires candidates to recall and describe specific critical incidents they have actually experienced in the past. Through systematic probing for concrete details, the method assesses demonstrated competencies, rather than relying on self-reported claims or idealized self-presentations. BEI is grounded in two core principles: First, past behavior is considered the most reliable predictor of future performance. How an individual has acted in similar situations historically strongly indicates how they are likely to act in the future. Second, the focus is on observable, specific actions within real events, not on stated attitudes or theoretical opinions. For instance, instead of asking, "What qualities do you think make a good leader?" a BEI interviewer would ask, "Describe a specific challenging situation you faced while leading a team. What exact steps did you take at the time?" Typically, a classic competency model development process based on BEI involves: 
\begin{itemize}
    \item Selecting criterion samples: defining high performers and average performers in the target position;
    \item Collecting data: using BEI methods on both groups of samples;
    \item Analyzing data: performing qualitative analysis on the collected text to extract factors that differentiate high performers from average performers, and establishing a competency model;
    \item Validating the model: using methods such as scale development and assessment centers to validate the model's effectiveness; 
    \item Applying the model: implementing the competency model across various stages of talent acquisition, development, and retention.
\end{itemize} 


The third step (data analysis) is the core of the whole process. In this stage, human experts meticulously dissect and compare the events gathered from interviews to ultimately identify the competencies required for the position. Typically, the expert focuses on what the individual did in the event, as well as their underlying thought process. Although this expert-based data analysis approach has been used for decades, it still has several significant limitations.

\begin{enumerate}
    \item {\bf High costs and lengthy modeling timelines}. Many companies lack the internal capability to conduct competency modeling and, therefore, often need to invest heavily in hiring external consulting firms. The expert teams from these firms typically require weeks or even months to analyze BEI data. As a result, the high cost and time demands limit the scalability of the competency modeling process across multiple job roles.
    \item {\bf The Quantity-Quality Paradox}. While acquiring more interview data and employing diverse competency libraries can theoretically enhance the quality and robustness of the model, it simultaneously significantly increases the pressure of information processing and the difficulty of comparative analysis. Consequently, the traditional modeling process cannot be feasibly improved through intuitive methods alone.
    \item {\bf Low objectivity and reproducibility.} Expert-driven modeling heavily relies on domain expertise and is highly subjective. Different expert teams may produce inconsistent results. Moreover, the data analysis process often involves conflicting opinions that require multiple rounds of group discussion to reach consensus. Consequently, the resulting models tend to lack reproducibility — even the same expert team may not arrive at identical results in repeated analyses.
    \item {\bf Difficulty in validation}. Competency models require validation to ensure their accuracy, typically through scale development and assessment centers. Scale development involves designing competency questionnaires to evaluate individuals in the same role (excluding those interviewed). Assessment centers evaluate competencies through written tests, interviews, leaderless group discussions, and situational exercises. Both methods assess the alignment between evaluation results and the competency model to determine validity. However, they are time-consuming and costly, and due to a lack of in-house expertise, organizations may skip the validation step altogether.
    \item {\bf Limited flexibility}. Results derived from expert teams are difficult to adjust based on key settings, such as the choice of competency library, the target level of modeling, or the number of key competencies.
\end{enumerate}
To this end, we design a new \underline{co}mpetency modeling process with \underline{LLM}s, namely {\bf CoLLM}, aiming for greater efficiency, improved objectivity and reproducibility, and enhanced flexibility. In addition, we can apply offline evaluation metrics to our proposed process, which facilitates comparisons across different designs, thereby complementing existing model validation approaches.
\subsection{New Process with LLMs}
\begin{figure}[!htbp]
\centering
\includegraphics[width=1\linewidth]{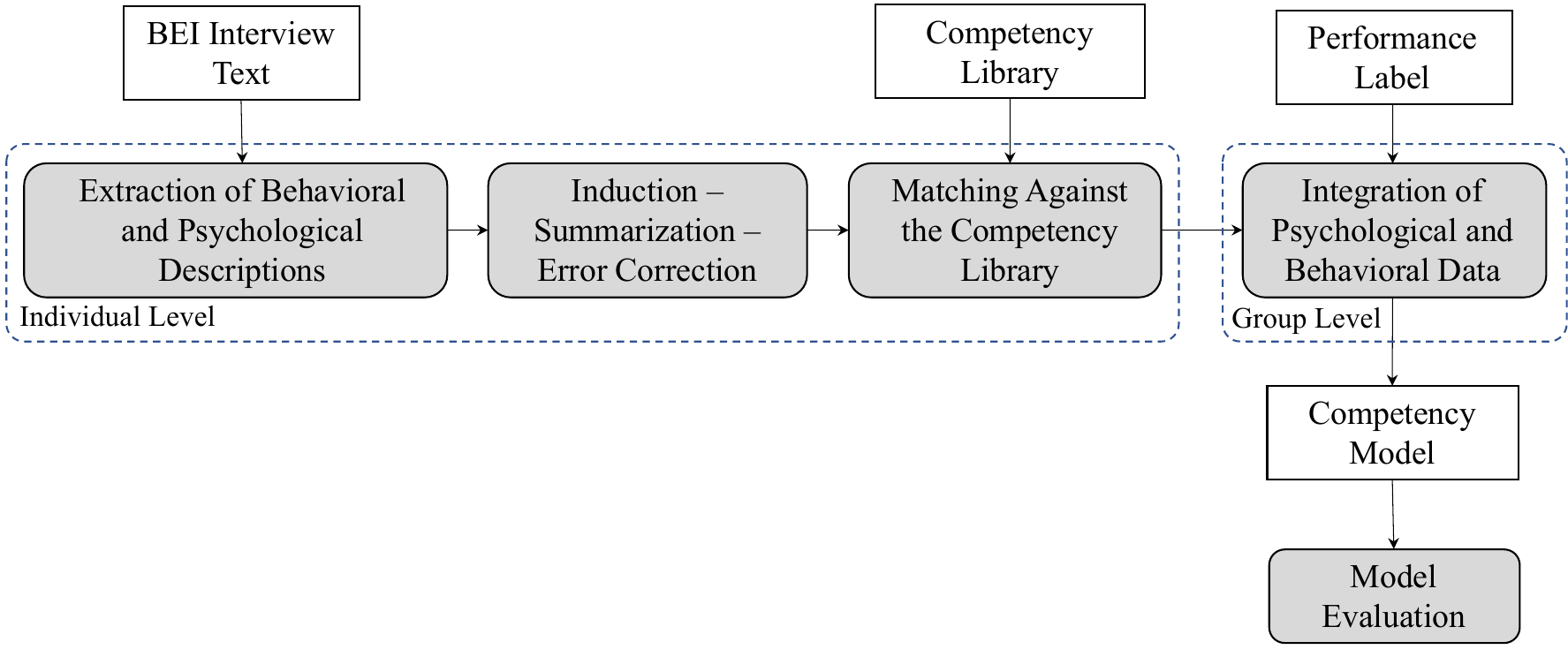}
\caption{The Workflow of CoLLM} 
\label{process}
\end{figure}
Figure~\ref{process} shows the workflow of our proposed method. Let $\mathcal{C}=\{c_1,\cdots,c_N\}$ be the set of participants in the target position with $N_1$ samples in the high-performance group and $N_2$ samples in the average-performance group. Each participant $c_i\in\mathcal{C}$ has a corresponding text record of the BEI $r_{c_i}\in\mathcal{R}$. We divide the set of participants $\mathcal{C}$ into the high-performance set $\mathcal{C}^+ =\{c_1^+,\cdots,c_{N_1}^+\}$ and the average-performance set $\mathcal{C}^-=\{c_1^-,\cdots,c_{N_2}^-\}$. The competency library is a structured collection of $L$ pre-defined competencies that are considered important for success within an organization, denoted as $\mathcal{M}=\{\left(m_1,T_1\right),\cdots,\left(m_L,T_L\right)\}$, where $m_i$ is the $i$-th competency item and $T_i$ is the textual description of $m_i$. Given the competency library, a competency model of participant $c$ is a set of scores $\boldsymbol{s}_{c}=\{s_1^c,s_2^c,\ldots,s_L^c\}$ on each competency item. We omit the subscripts or superscripts hereafter when there is no ambiguity.

CoLLM operates at two levels. (1) Individual level: For each participant, we extract behavioral and psychological descriptions from their interview transcript, and derive competency scores along these two dimensions, denoted as $\boldsymbol{s}^p$ and $\boldsymbol{s}^b$. (2) Group level: Based on the distribution of competencies in the high-performance group and the average-performance group, we learn a weight $\alpha$ to aggregate the behavioral and psychological scores. This yields group-level competency scores, denoted as $\boldsymbol{S}^+$ and $\boldsymbol{S}^-$.

\begin{figure}[!htbp]
\centering
\includegraphics[width=1\linewidth]{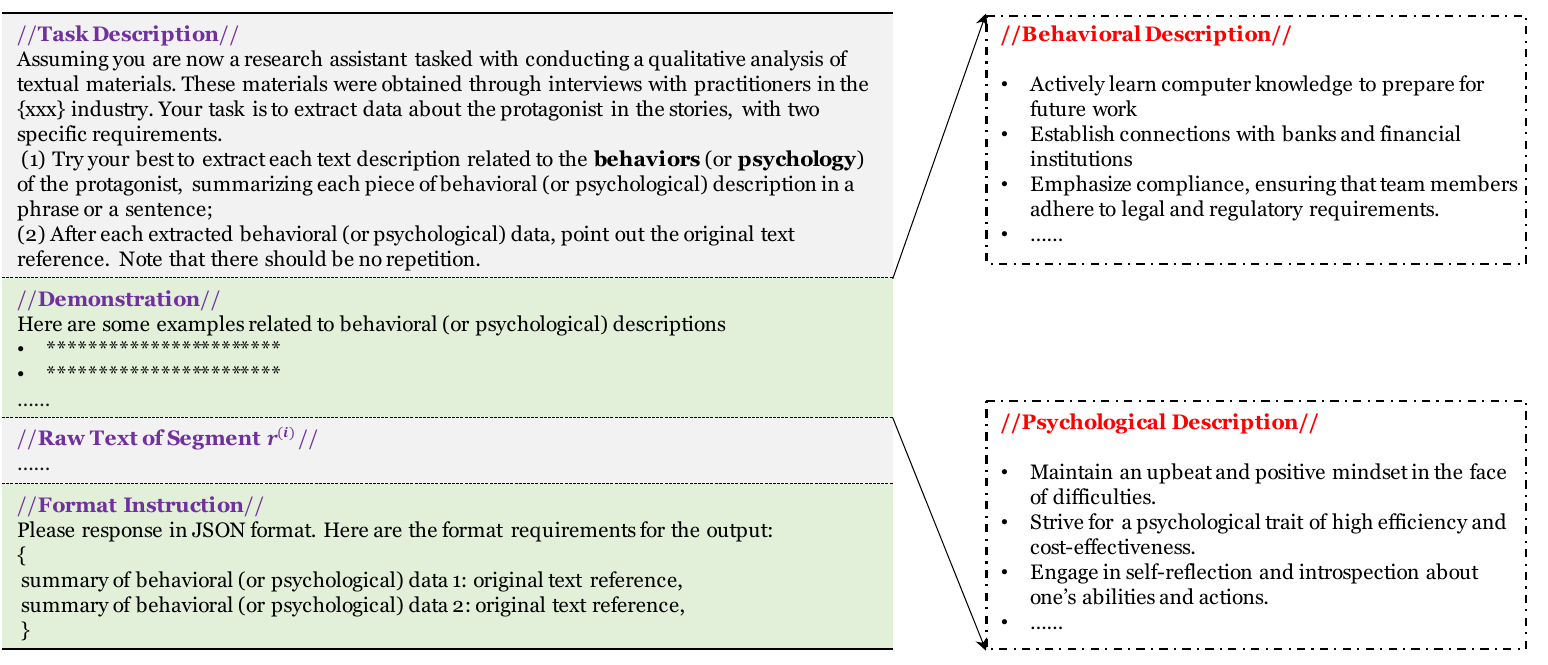}
\caption{Prompt Template for Extraction of Behavioral and Psychological Descriptions} 
\label{fig1}
\end{figure}

\paragraph{Individual Level.} During the BEI, participants are encouraged to describe several specific situations (i.e., events) they have encountered in the past, the actions they took, and the outcomes of those actions. We can easily divide the interview transcript $r$ of $c$ into $K$ distinct segments according to the different events described by the candidate, i.e., $r=\{r^{\left(1\right)},r^{\left(2\right)},\cdots,r^{\left(K\right)}\}$, where the superscript denotes the index of the event. In general, the number of segments $K$ is less than $5$.  Then we leverage the in-context learning (ICL) method, first proposed with GPT-3 \citep{Brown2020}, to extract descriptions related to behaviors and psychology from each segment $r^{\left(K\right)}$ through prompting a well-trained LLM. The prompt template is displayed in Figure \ref{fig1}, including a task definition and a few human-written task samples as demonstrations. 

To avoid omissions, we employ three different temperature parameters\footnote{The temperature parameter in LLMs is a hyperparameter that controls the randomness or creativity of the generated text. When the temperature is above 1, the model will generate more diverse and creative text. A temperature of 0 is equivalent to greedy decoding, leading to very repetitive and deterministic text.} to obtain multiple results.
\begin{equation}
    o_{b,\tau_i}^{\left(j\right)}=\texttt{ICL}\left(r^{\left(j\right)}, \tau_i, t_b\right),  o_{p,\tau_i}^{\left(j\right)}=\texttt{ICL}\left(r^{\left(j\right)}, \tau_i, t_p\right),  \tau_i\in\{\tau_1, \tau_2, \tau_3\}, j= 1,\cdots K
\end{equation}
where $t_b$ and $t_p$ are prompt templates for behavioral and psychological analysis respectively, $\tau_i$ is the temperature parameter, and $o_{b,\tau_i}^{\left(j\right)}$, $o_{p{,\tau}_i}^{\left(j\right)}$ are extracted behavioral and psychological descriptions in the segment  $r^{\left(j\right)}$. In this step, we control the generation seed of the LLM to ensure reproducibility. Behavioral and psychological descriptions are at the core of competency modeling, and any errors or omissions could potentially affect the outcomes. We introduce an LLM-based review step to integrate results from different temperatures while also filtering out incorrect descriptions. The prompt templates in this step are roughly the same as those in Figure \ref{fig1}, with a slightly different task description.
\begin{equation*}
{\tilde{o}}_b^{\left(j\right)}=\texttt{LLMReview}\left(o_{b,\tau_1}^{\left(j\right)},o_{b,\tau_2}^{\left(j\right)},\ o_{b,\tau_3}^{\left(j\right)}\right),\quad {\tilde{o}}_p^{\left(j\right)}=\texttt{LLMReview}\Big (o_{p,\tau_1}^{\left(j\right)},o_{p,\tau_2}^{\left(j\right)},\ o_{p,\tau_3}^{\left(j\right)}\Big )
\end{equation*}
After that, we transform the behavioral and psychological data of each event to document embeddings, i.e.,  $$\boldsymbol{o}_b =\texttt{emb}({\tilde{o}}_b^{\left(1\right)},{\tilde{o}}_b^{\left(2\right)},\cdots,{\tilde{o}}_b^{\left(K\right)}), \boldsymbol{o}_p =\texttt{emb}({\tilde{o}}_p^{\left(1\right)},{\tilde{o}}_p^{\left(2\right)},\cdots,{\tilde{o}}_p^{\left(K\right)})$$ Similarly, we can represent each competency item $m_i$ of the library in the same latent semantic space by mapping the textual description $T_i$ to an embedding vector, i.e., $\boldsymbol{t}_i=\texttt{emb}(T_i), i=1, \cdots, L$. Then we calculate two scores on each competence item $m_i$ through the behavioral and psychological data.
\begin{equation}\label{eq2}
    s_i^b=\cos<\boldsymbol{t}_i,\boldsymbol{o}_b >=\frac{\boldsymbol{t}_i^T\boldsymbol{o}_b}{\left|\boldsymbol{t}_i\right|\cdot\left|\boldsymbol{o}_b \right|},\quad  s_i^p=\cos<\boldsymbol{t}_i,\boldsymbol{o}_p>,\quad  i=1,\cdots,L
\end{equation}

\paragraph{Group Level.} Given $\boldsymbol{s}^b=[s_1^b,\ \cdots, s_L^b]$ and ${\boldsymbol{s}^p=[s}_1^p,\ \cdots, s_L^p]$ for each participant, we average scores of individuals within the high-performance and average-performance groups to get group-level results, i.e., $\boldsymbol{S}^{b,+}, \boldsymbol{S}^{p,+}$ and $\boldsymbol{S}^{b,-}, \boldsymbol{S}^{p,-}$. To obtain the final competency model, we need to combine the results derived from behavioral and psychological data. One straightforward method is to simply add them together, assuming that psychological and behavioral data play the same role in the competency model. However, for different organizations, or even different positions within the same organization, the weight of psychological and behavioral data in the competency model may not be the same. For this reason, we introduce a learnable weight, $\alpha$, to integrate group-level behavioral and psychological results.
\begin{equation}\label{eq3}
\boldsymbol{S}^+=\boldsymbol{S}^{b+}+\alpha\cdot\boldsymbol{S}^{p,+},\quad \boldsymbol{S}^-=\boldsymbol{S}^{b,-}+\alpha\cdot\boldsymbol{S}^{p,-}
\end{equation}
Theoretically, a good $\alpha$ should result in a significant difference in competency scores between the high-performance group and the average-performance group. We design a loss function based on the triplet loss used in face recognition \citep{Schroff2015} to learn $\alpha$. Each training sample is a triplet $(c,c',c'')$ consisting of a randomly selected participant $c\in \mathcal{C}$, an individual $c'$ from the same performance group as $c$, and a participant $c''$ from the other group. The similarity of competency scores weighted by $\alpha$ within the same performance group is desired to be higher than that between different groups. Therefore, we define the following loss function.
\begin{equation}\label{eq4}
l\left(c,c^\prime,c^{\prime\prime};\alpha\right)=\cos<\boldsymbol{s}_c^b+\alpha\boldsymbol{s}_c^p,\ {\ \boldsymbol{s}}_{c^{\prime\prime}}^b+\alpha\boldsymbol{s}_{c^{\prime\prime}}^p>\ -\ \cos<\boldsymbol{s}_c^b+\alpha\boldsymbol{s}_c^p,\ {\ \boldsymbol{s}}_{c^\prime}^b+\alpha\boldsymbol{s}_{c^\prime}^p>
\end{equation}
Subsequently, we leverage stochastic gradient descent to optimize the loss function and get the optimized weighting parameter $\alpha$. We obtain the final competency models of two performance groups by plugging $\alpha$ into Eq. (\ref{eq3}). Now we can rank each item in the competency library based on the scores difference between the two performance groups, i.e., $\boldsymbol{S}^+-\boldsymbol{S}^-$, thereby identifying the critical competencies.
\subsection{Offline Evaluation}\label{section3.3}
Existing methods for validating competency models typically require large-scale competency assessments of other individuals in the same role (e.g., through questionnaires or assessment centers). We refer to these methods relying on the collection of new data as online evaluation methods, which often entail high costs and long assessment cycles. As a result, organizations may skip rigorous validation steps in practice. In contrast, this subsection focuses on offline evaluation methods without the collection of new data.

First, we randomly select $N_{\mathrm{test}}$ participants (around 20\%–25\%) from $\mathcal{C}$ as a test set $\mathcal{C}_{\mathrm{test}}$, with the remaining participants forming the training set. On the training set, we learn $\alpha$ to compute the competency scores for the two performance groups, denoted as  $\boldsymbol{S}_{\mathrm{train}}^+$ and $\boldsymbol{S}_{\mathrm{train}}^-$. Next, we identify the top-$Q$ competency items with the largest differences in $\boldsymbol{S}_{\mathrm{train}}^+-\boldsymbol{S}_{\mathrm{train}}^-$, and treat them as key competencies. We then apply the same process and the learned $\alpha$ to $\mathcal{C}_{\mathrm{test}}$ to compute each participant’s average score across the $Q$ key competencies. Based on these average scores, we rank all participants in $\mathcal{C}_{\mathrm{test}}$, resulting in a ranking $Y=\left(y_1,y_2,\ldots,y_{N_{\mathrm{test}}}\right)$, where $y_i\in\mathcal{C}_{\mathrm{test}}$ is the participant ranked at $i$-th position in $Y$. Similarly, we rank the same participants based on their actual performance to obtain $Z=\left(z_1,z_2,\ldots,z_{N_{\mathrm{test}}}\right), z_i\in\mathcal{C}_{\mathrm{test}}$. We use Spearman’s rank correlation \citep{Spearman1904} to quantify the similarity between $Y$ and $Z$, thereby reflecting the extent to which the key competencies extracted from the training set align with the actual performance.
\begin{equation}
    \rho=1-\frac{6\sum_{c\in\mathcal{C}_{\mathrm{test}}}\left({\rm rank}_Y\left(c\right)-{\rm rank}_Z\left(c\right)\right)^2}{N_{\mathrm{test}}\left({N_{\mathrm{test}}}^2-1\right)}
\end{equation}
where ${\rm rank}_Y\left(c\right)$ and ${\rm rank}_Z\left(c\right)$ denote the rank of $c$ in $Y$ and $Z$, respectively. 

In practice, a full ranking of participant performance might be unavailable; instead, each candidate is typically assigned a binary performance label. In this setting, we treat the average key-competency score as a continuous prediction and use the Area Under the ROC Curve (AUC) as an additional evaluation metric.

\section{Experiments}
In this section, we conduct extensive experiments to answer the following questions.
\begin{itemize}
    \item Can our CoLLM framework produce a coherent and interpretable competency model for a given target role?
    
    We begin by presenting a full modeling example to illustrate how behavioral and psychological evidence is transformed into competency scores and how key competencies emerge.
    \item How to determine the number of key competencies?
    
    We evaluate competency models constructed under different numbers of key competencies using the evaluation method described in Section \ref{section3.3}, and identify the number that yields the best overall performance.
    \item How robust is the modeling process when we replace the underlying LLM or compare against human expert coders?

    We evaluate the generalizability of our framework by changing the backend LLMs. Besides, we invite human experts to manually extract behavioral and psychological descriptions from interview transcripts. This allows us to assess the reliability of substituting human coding with LLM-based extraction.
    \item Are the identified key competencies consistent across different competency libraries?

    We replace the target competency framework with alternative libraries and replicate the full pipeline. 
\end{itemize}

\subsection{Data Collection and Settings}
To evaluate the CoLLM process, we collaborate with a company in the software outsourcing industry. We choose the team leaders (TLs) as the target position in competency modeling. These TLs serve as the on-site leaders for outsourced employees at the client company, responsible for coordinating task delivery, team development, and communication between the client and the outsourcing team. There are hundreds of TLs in this company and we conduct behavioral event interviews (BEI) with 40 TLs in a month, with 18 in the high-performance group and 22 in the average-performance group. The criterion for grouping is the performance of each manager in the previous 3 months. Each interview lasts about 2 hours, and all participants mention at least three events that have played a significant role in their careers. 

We use the “67 Lominger Competencies (Fifth Edition)” \citep{LombardoEichinger2010} as the competency library, which has already been widely adopted by the company. The library contains three levels with 67 competencies at the third level, 20 clusters at the second level, and 6 factors at the top level. Some examples of the library are displayed in Table \ref{tab1}. Each cluster and competency has a short text description, i.e., $T_i$ in our process. We focus on the 20 clusters at the second level as the modeling targets. Due to copyright restrictions, we do not provide a detailed discussion of the specific meanings of these clusters. However, Appendix \ref{appendix_a} lists the names and brief descriptions of all 20 clusters. Interested readers may refer to \citet{LombardoEichinger2010} for a comprehensive explanation. We follow the default numbering in the competency library, using “A” to “T” to represent the 20 clusters of competencies. Consistent with the traditional practice, we first set the number of key competencies to one-third of the total competencies, which yields $Q = 7$ in our context. We then further explore how to determine the optimal value of $Q$ in Section \ref{sec4.3}. We select Qwen2.5-Max \citep{Yang2024} as the primary LLM in the CoLLM process with $\tau_1=0,\tau_2=0.5$, and $\tau_3=1$. The random seed is set to 2025 to ensure reproducibility.

\begin{table}[!htbp]
  \centering
    \caption{Structure of the Competency Library and Some Examples}
    \label{tab1}
    \begin{tabular}{L{3.5cm}|L{6cm}|L{6cm}}
    \toprule
    \midrule
    Top-level (Factors)  & Second-level (Clusters) & Third-level (Competencies)  \\
    \midrule
    \multirow{3}{*}{I: {\em Strategy Skills}}  & {A: {\em Understanding the Business}\begin{itemize}
         \item Is familiar with the business;
        \item Master the key technologies and skills required in work;
        \item Knows operational programs and their general operation;
        \item Can learn new methods and techniques;
     \end{itemize}}  &  5. {\em Business Acumen} \begin{itemize}
         \item Is familiar with the business;
        \item Master the key technologies and skills required in work;
        \item Knows operational programs and their general operation;
        \item Is able to learn new methods and techniques;
     \end{itemize}\\
     \cline{3-3}
     & & 24. {\em Functional/Technical Skills }\\
     \cline{3-3}
     & & 61. {\em Technical Learning} \\
    \bottomrule
    \end{tabular}%
\end{table}%
\subsection{Competency Modeling Results}
\begin{figure}[!htbp]
  \centering
  \subfigure[]{\label{fig2a}
  \includegraphics[width=0.44\linewidth]{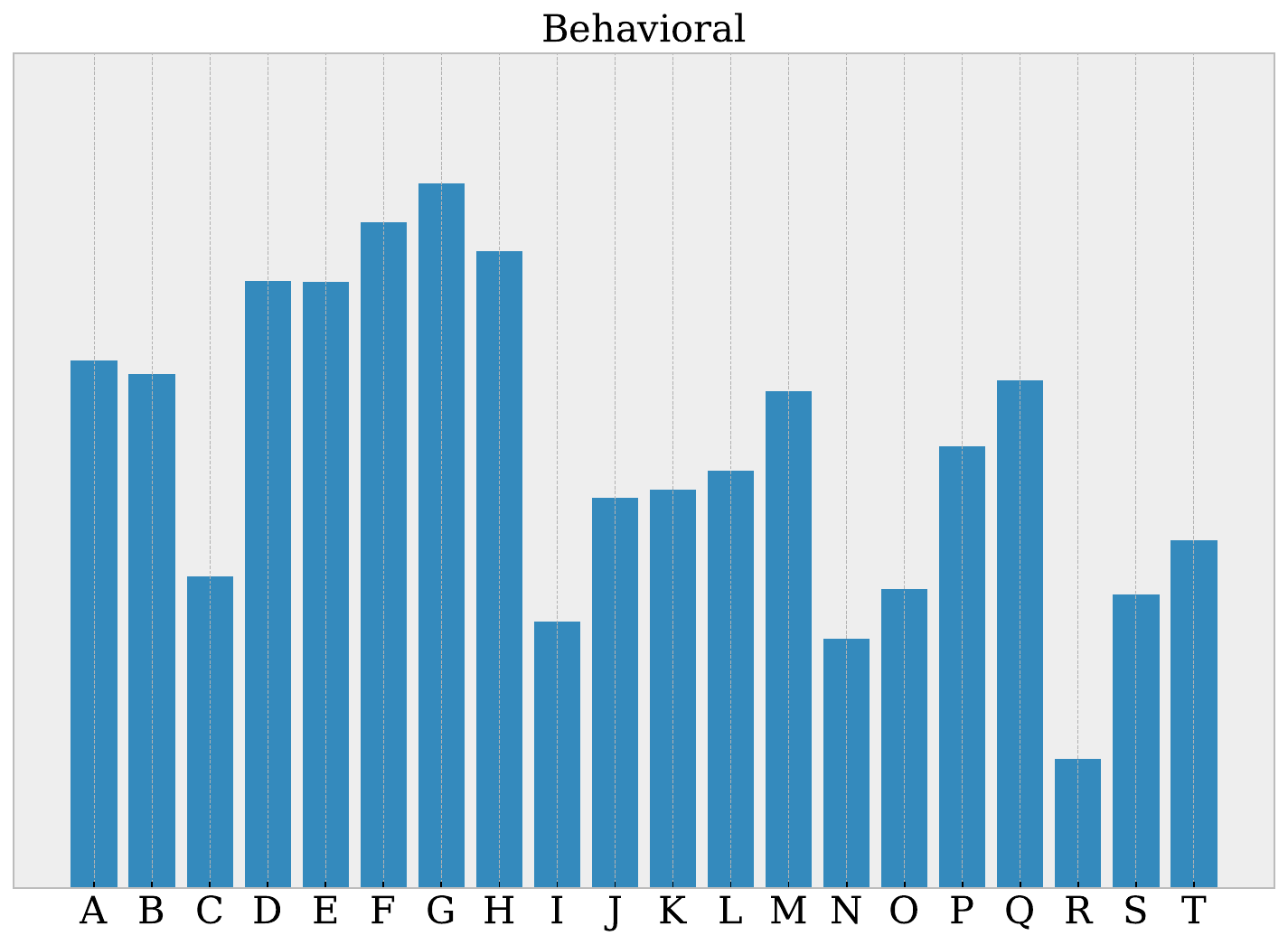}
  }
  \subfigure[]{\label{fig2b}
  \includegraphics[width=0.44\linewidth]{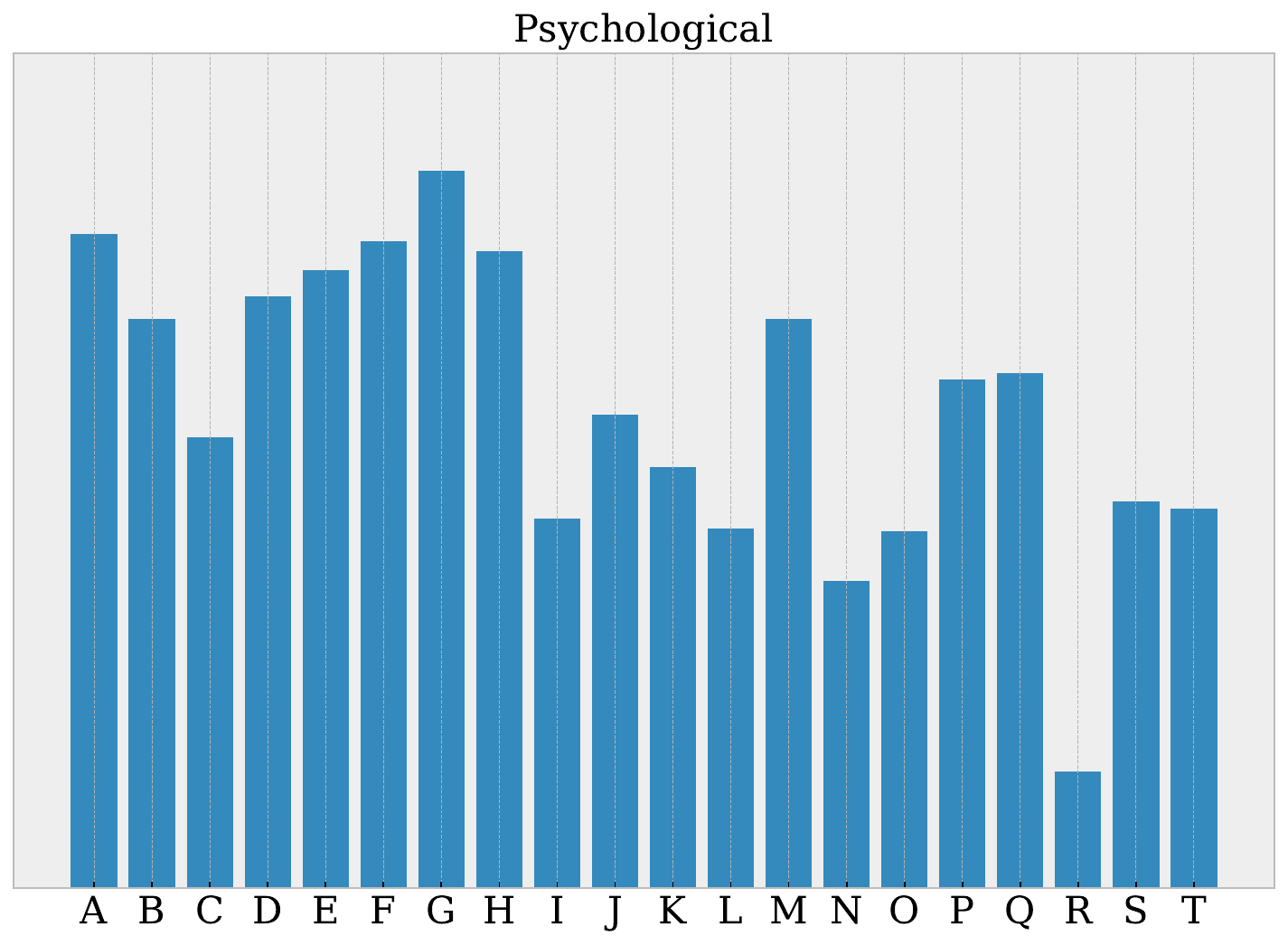}
 }
  \caption{Scores on Each Competence Item Derived from Behavioral (Left) and Psychological (Right) Descriptions of a Participant}
  \label{fig2}
\end{figure}

\begin{figure}[!htbp]
\centering
\includegraphics[width=0.7\linewidth]{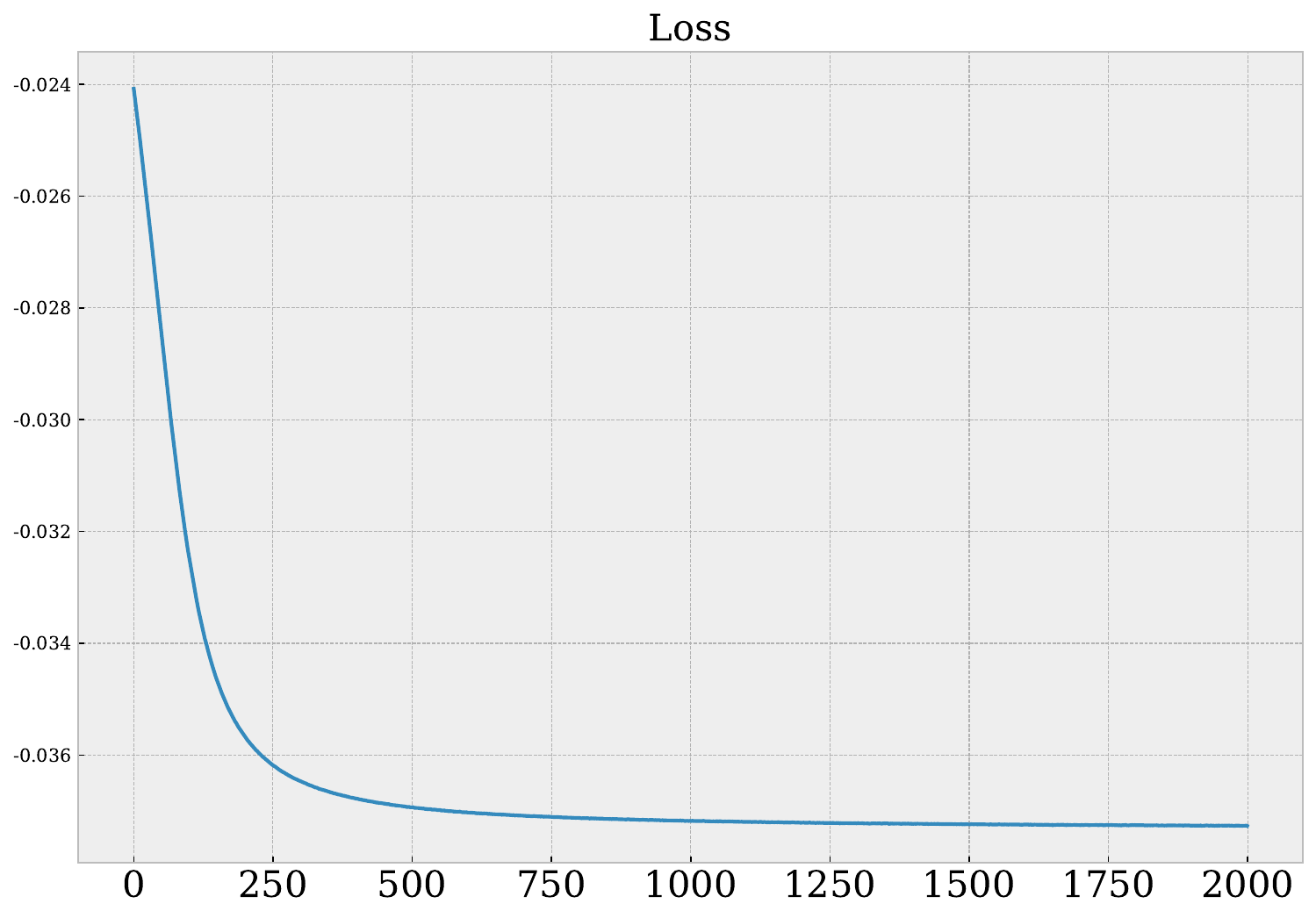}
\caption{Loss for Each Epoch} 
\label{loss}
\end{figure}

In this subsection, all 40 participants are included for modeling. Figure \ref{fig2} shows the $\boldsymbol{s}^b$ and $\boldsymbol{s}^p$ of a selected participant. The top-7 clusters of competencies derived from his behavioral data are L/A/G/F/B/H/T, and those obtained from his psychological data are B/G/T/L/H/F/P. After getting $\boldsymbol{s}^b$ and $\boldsymbol{s}^p$ for each participant, we construct 400 triplets from our training set and use AdamW \citep{LoshchilovHutter2018} to optimize Eq. (\ref{eq4}) for 2000 epochs. Figure~\ref{loss} illustrates the variation of the loss during the training process. The optimized $\alpha$ is 12.23, indicating that for the TLs of this company, the psychological data in their competencies is much more important than the behavioral data. Traditional processes of competency modeling are not capable of analyzing or revealing this finding. 

\begin{figure}[!htbp]
\centering
\includegraphics[width=0.7\linewidth]{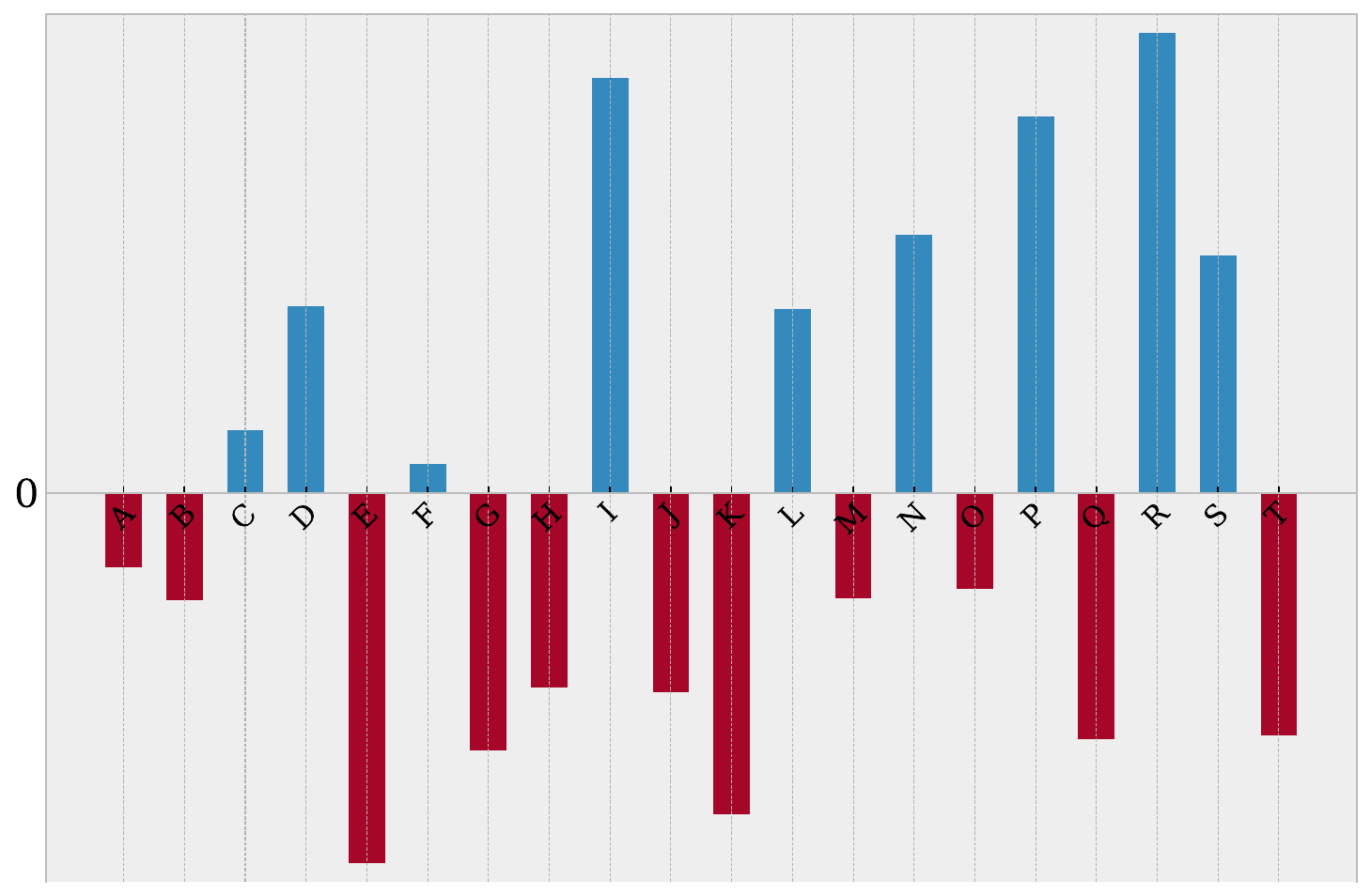}
\caption{Differences in Scores between the High-performance Group and the Average-performance Group} 
\label{fig3}
\end{figure}

Plugging $\alpha=12.23$ into Eq. (\ref{eq3}) yields the group-level results, shown in Figure \ref{fig3}. We summarize that the seven most critical clusters of competencies are R/I/P/N/S/D/L.  This suggests that for the branch managers of this company, the following seven capabilities are very important: acting with honor and character, making tough people calls, managing diverse relationships, relating skills, Being Open and Receptive, keeping on point, and communicating effectively.

Besides, we invite two experts to analyze the participant presented in Figure \ref{fig2}. They rigorously follow the traditional process to code the BEI data and determine which clusters of competencies the participant aligns with. The top-5 competencies identified by experts are B/H/L/P/T. We leverage the optimized $\alpha$ to integrate $\boldsymbol{s}^b$ and $\boldsymbol{s}^p$ of this participant, and the seven most significant clusters are B/G/L/T/H/F/P, containing all competencies found by human experts. This demonstrates the validity and rationality of using LLMs for competency analysis.

\subsection{Determine the Number of Key Competencies}\label{sec4.3}

In practice, the number of key competencies \( Q \) is typically selected within a relatively narrow range, often between 5 and 10. In traditional expert-driven competency modeling, this choice largely depends on the collective judgment and experience of the expert team. While such an approach benefits from domain expertise, it suffers from several limitations. First, the selection of \( Q \) is often subjective and lacks a transparent quantitative justification. Second, different expert teams may arrive at different values of \( Q \), leading to limited reproducibility. Third, without systematic validation, the chosen $Q$ may not optimally distinguish high performers from average performers.

In contrast, our LLM-based workflow enables rapid and repeated construction of competency models under different configurations. Leveraging the offline evaluation framework introduced in Section \ref{section3.3}, we can systematically vary \( Q \) within the range of 5 to 10 and identify the value that yields the best predictive performance. This data-driven procedure transforms the selection of \( Q \) from a heuristic decision into an empirically grounded model selection problem.

Specifically, we use a four-fold cross-validation procedure. The dataset is evenly divided into four subsets. In each round, one subset serves as the test set, and the remaining three are used to build the competency model by the CoLLM pipeline. We then select the top-\( Q \) competencies with the largest performance-group differences and compute each participant’s average score on these competencies in the test set. The resulting ranking is compared with their actual performance. The evaluation metric in Section~\ref{section3.3} is calculated for each fold, and the four results are averaged.

This procedure allows us to examine how different values of \( Q \) influence the model’s ability to distinguish performance levels and to identify the optimal number of key competencies.

\begin{figure}[!htbp]
  \centering
  \subfigure[]{\label{fig2a}
  \includegraphics[width=0.44\linewidth]{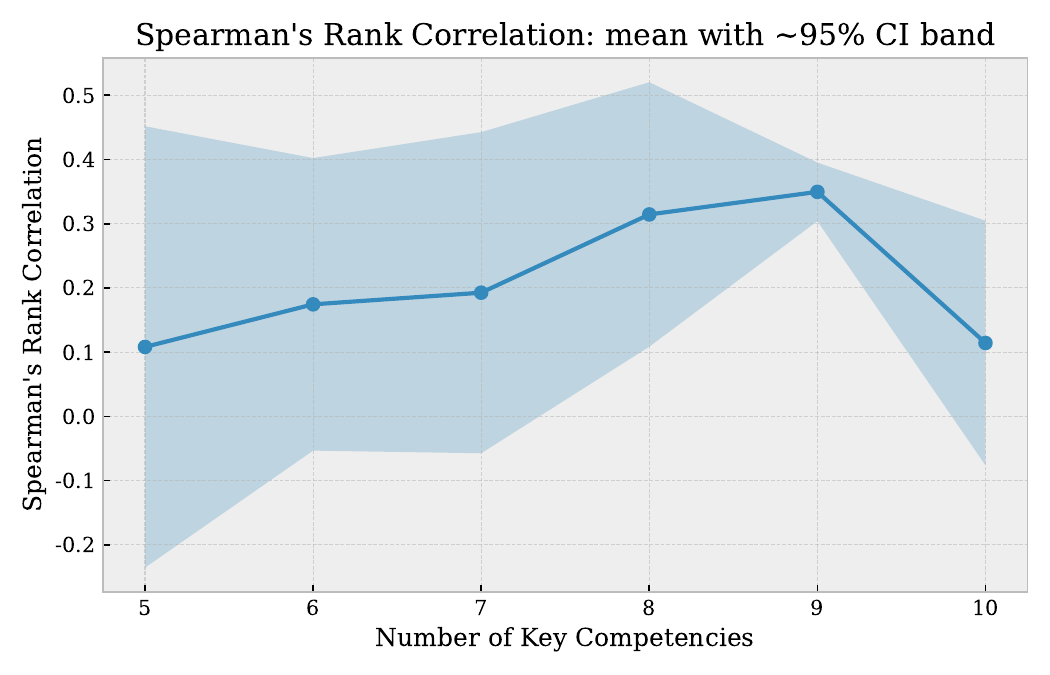}
  }
  \quad 
  \subfigure[]{\label{fig2b}
  \includegraphics[width=0.44\linewidth]{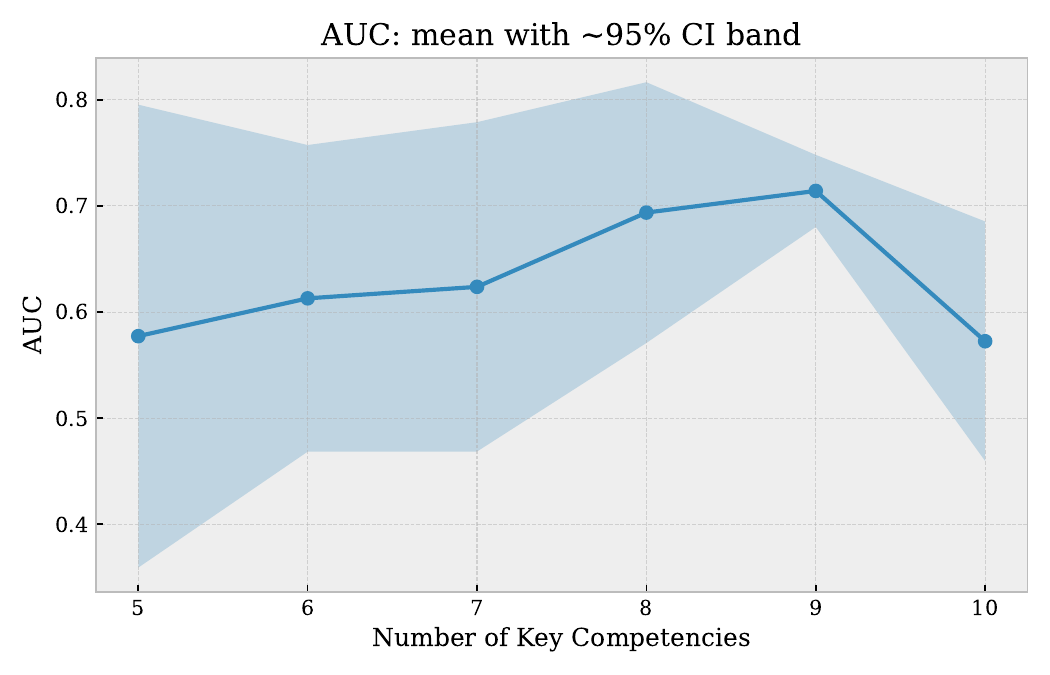}
 }
  \caption{Offline Evaluation Results with Different $Q$}
  \label{fig6}
\end{figure}

Figure \ref{fig6} presents the offline evaluation results. As shown in the figure, the competency model generated by CoLLM achieves the best performance when \( Q = 9 \). On the test set, the AUC between the ranking based on key competencies and the actual performance ranking exceeds 0.7, and the Spearman rank correlation coefficient is greater than 0.3. These results indicate that the selected key competencies exhibit strong discriminative ability and a meaningful positive association with actual performance, suggesting good predictive validity of the model. 

Based on the results above, we set \( Q = 9 \) and use the full sample to construct the final competency model. Compared with the results reported in Subsection 4.2, two additional clusters, C and F, are included among the key competencies. These correspond to creating the new and different and getting work done through others, respectively.

\subsection{Ablation and Robustness Analysis}
Using the same four-fold cross-validation procedure described in Subsection~\ref{sec4.3}, we compare the performance of CoLLM under different backend LLMs. Furthermore, we conduct additional analyses by replacing or removing specific modules within CoLLM. The following variants are considered for comparison:
\begin{enumerate}
    \item CoLLM-GLM and CoLLM-Doubao: Variants of the CoLLM framework using different LLMs, specifically GLM-4-plus and Doubao-1.5-pro, respectively.
    \item CoLLM$^-$: A simplified version of the CoLLM where we remove the step of learning $\alpha$ and set $\alpha=1$, treating behavioral and psychological data as equally important.
    \item CoLLM-expert: three competency modeling experts manually extract behavioral and psychological descriptions from the interview transcripts, while all other steps remain unchanged.
\end{enumerate}

\begin{table}[!htbp]
  \centering
    \caption{Evaluation Results of Different Variants}
    \label{tab2}
    \begin{tabular}{L{3cm}|L{2cm}|L{2cm}|L{2cm}|L{4cm}}
    \toprule
    \midrule
    Method & Optimal $Q$ & AUC & $\rho$ & Key Competencies  \\
    \midrule
    CoLLM & 9 & 0.715 & 0.350 & R, I, P, N, S, D, L, C, F \\
    \midrule
    CoLLM-GLM & 8 & 0.707 & 0.332 & R, P, I, C, D, N, L, S  \\
    CoLLM-Doubao & 8 & 0.711 & 0.3467 & P, I, D, C, R, L, S, F\\
    CoLLM$^-$ & 7 & 0.6571 & 0.267 & I, P, R, N, S, M, T \\
    CoLLM-expert & 6 & 0.721 & 0.351 & N, R, S, P, T, F \\
    \bottomrule
    \end{tabular}%
\end{table}%

Table \ref{tab2} shows the evaluation results. First, CoLLM achieves strong and stable performance across different backend LLMs, with only slight fluctuations in AUC and \( \rho \) when replacing the underlying model. This suggests that the overall framework is robust to the choice of LLM. Second, removing the weight-learning step (CoLLM\(^{-}\)) leads to a noticeable performance drop, indicating that dynamically learning the relative importance of behavioral and psychological data is critical for model effectiveness. Third, the CoLLM-expert variant achieves performance comparable to the original CoLLM, and even slightly higher AUC, suggesting that LLM-based extraction can effectively approximate expert-level coding. 

Beyond performance metrics, the identified key competencies are also highly consistent across variants. Most backend LLMs produce overlapping sets of competencies, with core clusters such as R, P, N, S, and L appearing repeatedly. Importantly, most of the competencies identified by CoLLM-expert are also captured by CoLLM , indicating strong structural agreement between LLM-based extraction and human coding.

Overall, the results demonstrate both the robustness of the framework and the importance of its key design components.

\subsection{Cross-library Robustness}
Traditional expert-driven competency modeling needs to fix the target library at the beginning. Once the process begins, changing the framework requires much extra work, such as re-coding interview data and re-aggregating results. In contrast, the competency library functions as a plug-and-play module in the CoLLM framework. Our pipeline allows the target library to be replaced without altering the upstream extraction and scoring procedures, thereby providing significantly greater flexibility.

In this subsection, we evaluate the robustness of our framework by replacing the original competency library with alternative libraries. First, we adopt an alternative competency framework developed by Korn Ferry. This library is single-layered and consists of 38 competencies organized into four dimensions: Thinking, Results, People, and Self. Table \ref{tab3} presents several illustrative examples. The 38 competency items in the Kon Ferry library are indexed from 1 to 38.

\begin{table}[!htbp]
  \centering
    \caption{Some Examples of the Alternative Competency Library}
    \label{tab3}
    \begin{tabular}{L{2cm}|L{3.5cm}|L{3cm}|L{6cm}}
    \toprule
    \midrule
    Number & Dimension  & Competency & Definition  \\
    \midrule
    1 & Thinking & Business Insight & Applies knowledge of business and the marketplace to advance the organization’s goals.\\  
    \midrule
    11 & Results & Action-Oriented & Takes on new opportunities and tough challenges with a sense of urgency, high energy, and enthusiasm. \\
    \bottomrule
    \end{tabular}%
\end{table}%

In addition, we conduct modeling using the third level of the 67 Lominger Competencies, treating all 67 individual competency items as modeling targets instead of the 20 second-level clusters. The 67 competency items are indexed from 1 to 67. For simplification, we refer to the two levels of the 67 Lominger Competencies as the Longmier second level and the Longmier third level, respectively. 

For both alternative libraries, we compare CoLLM and CoLLM-expert. We first follow the four-fold cross-validation procedure described in Section \ref{sec4.3} to obtain evaluation results and determine the optimal value of $Q$. We then use the full sample to construct the final competency model under each library. 

\begin{table}[!htbp]
  \centering
    \caption{Results of Different Competency Libraries}
    \label{tab4}
    \begin{tabular}{L{3cm}|L{3cm}|L{1cm}|L{1.5cm}|L{1.5cm}|L{4cm}}
    \toprule
    \midrule
    Method & Library & $Q$ & AUC & $\rho$ & Key Competencies  \\
    \midrule
     \multirow{3}{*}{CoLLM} & Longmier Second Level & 9 & 0.715 & 0.350 & R, I, P, N, S, D, L, C, F \\
     & Longmier Third Level & 8 & 0.691 & 0.293 & 37, 33, 25, 3, 15, 41, 67, 51\\
     & Kon Ferry & 9 & 0.725 & 0.349 & 20, 29, 2, 19, 21, 13, 26, 22, 24 \\
    \midrule
    \multirow{3}{2cm}{CoLLM-expert } & Longmier Second Level & 6 & 0.721 & 0.351 & N, R, S, P, T, F \\
    & Longmier Third Level & 9 & 0.502 & $-0.017$ & 25, 3, 37, 23, 26, 60, 44, 21, 29 \\
    & Kon Ferry & 5 & 0.501 & 0.001  & 26, 20, 23, 24, 22 \\
    \bottomrule
    \end{tabular}%
\end{table}%

Table \ref{tab4} reports the evaluation results across different competency libraries. First, CoLLM achieves consistently strong performance under all three libraries. The AUC remains around 0.69–0.73 and \( \rho \) around 0.29–0.35, indicating stable discriminative ability across alternative competency structures. In contrast, CoLLM-expert shows substantial performance variation. While it performs comparably to CoLLM under the Longmier second level (AUC = 0.721, \( \rho = 0.351 \)), its performance drops sharply under the Longmier third level and the Kon Ferry library (AUC $\approx$ 0.50 and \( \rho \) close to 0). This suggests that expert-based extraction is more sensitive to changes in library granularity and structure, whereas CoLLM generalizes more robustly across different competency frameworks.

Second, the two methods exhibit high structural consistency within the Longmier library. For example, the competencies identified by CoLLM at the third level (33, 25, 3, 15, 41, 67) correspond to clusters S, I, N, P, S, and L at the second level, which are also included in the second-level results. This hierarchical alignment indicates that the extracted competencies are conceptually coherent across levels. Furthermore, Table \ref{tab4} also reveals cross-library consistency. For instance, CoLLM identifies clusters I (Making Tough People Calls), L (Communicating Effectively), and F (Getting Work Done Through Others) at the Longmier second level. These clusters correspond closely to competency items 22 (Attracts Top Talent), 26 (Communicates Effectively), and 13 (Directs Work) in the Korn Ferry library. This cross-library alignment suggests that CoLLM captures underlying competency constructs rather than artifacts specific to a particular taxonomy, thereby demonstrating conceptual robustness beyond any single predefined competency structure.

Third, in the Kon Ferry library, the nine key competencies identified by CoLLM cover four out of the five competencies identified by CoLLM-expert, further confirming that the extraction process traditionally performed by human experts can be effectively replicated by LLMs.

These results highlight the adaptability of the proposed framework and its ability to preserve conceptual consistency under varying competency taxonomies.

\section{Conclusions and Future Directions}

Competency modeling has long been criticized for its limited rigor, high implementation costs, and heavy reliance on expert judgment. In this study, we propose a novel LLM-enhanced workflow that systematically restructures the traditional competency modeling process. By integrating LLMs into competency extraction, scoring, and model selection, our framework transforms what has historically been a largely subjective procedure into a more transparent, data-driven, and reproducible process.

Through extensive offline evaluations, we demonstrate that the proposed CoLLM framework achieves strong predictive validity, robust performance across alternative competency libraries, and high structural consistency across different levels of granularity. The framework not only approximates expert-level extraction but also introduces a fully data-driven optimization mechanism that allows the entire modeling process to determine the most suitable configuration, such as the optimal number of key competencies. These features significantly enhance modeling efficiency, reduce subjective bias, and substantially lower implementation costs. As a result, competency modeling becomes more scalable and accessible, particularly for small and medium-sized enterprises that may lack dedicated expert teams.

This study opens several avenues for future research. First, while our validation relies on cross-sectional performance differences, future work could incorporate longitudinal or lagged performance data to further examine the causal and predictive validity of the constructed competency models. Second, expanding the framework to larger and more diverse organizational samples would allow researchers to test its generalizability across industries and cultural contexts. Finally, investigating the interpretability and governance implications of LLM-assisted competency modeling represents an important direction, especially as organizations increasingly rely on AI-supported decision systems.


%
%
%



\bibliographystyle{informs2014} 
\bibliography{ref} 

\begin{APPENDICES}
\section{Competency Library}\label{appendix_a}
In this appendix, we provide a brief description of the competency items from the Lominger and Korn Ferry libraries that are referenced in Table~\ref{tab2} and Table~\ref{tab4}.

\subsection{67 Longmier Competencies}
\begin{itemize}
    \item \textbf{B}: Making Complex Decisions
        \begin{itemize}
            \item [] \textit{51}: Problem Solving
        \end{itemize}
    \item \textbf{C}: Creating the New and Different
    \item \textbf{D}: Keeping on Point
    \item \textbf{F}: Getting Work Done Through Others
    \item \textbf{I}: Making Tough People Calls
        \begin{itemize}
            \item [] \textit{25}: Hiring and Staffing
        \end{itemize}
    \item \textbf{L}: Communicating Effectively
        \begin{itemize}
            \item [] \textit{67}: Written Communications
        \end{itemize}
    \item \textbf{M}: Managing Up
    \item \textbf{N}: Relating Skills
        \begin{itemize}
            \item [] \textit{3}: Approachability
        \end{itemize}
    \item \textbf{P}: Managing Diverse Relationships
        \begin{itemize}
            \item [] \textit{15}: Customer Focus
            \item [] \textit{21}: Managing Diversity
            \item [] \textit{23}: Fairness to Direct Reports
        \end{itemize}
    \item \textbf{Q}: Inspiring Others
        \begin{itemize}
            \item [] \textit{37}: Negotiating 
            \item [] \textit{60}: Building Effective Teams
        \end{itemize}
    \item \textbf{R}: Acting With Honor and Character
        \begin{itemize}
            \item [] \textit{29}: Integrity and Trust
        \end{itemize}
    \item \textbf{S}: Being Open and Receptive
        \begin{itemize}
            \item [] \textit{26}: Humor 
            \item [] \textit{33}: Listening 
            \item [] \textit{41}: Patience
            \item [] \textit{44}: Personal Disclosure
        \end{itemize}
    \item \textbf{T}: Demonstrating Personal Flexibility
\end{itemize}

\subsection{Kon Ferry Library}
\begin{itemize}
    \item \textbf{2}: Customer Focus
    \item \textbf{13}: Directs Work
    \item \textbf{19}: Manages Conflicts
    \item \textbf{20}: Interpersonal Savvy
    \item \textbf{21}: Builds Networks
    \item \textbf{22}: Attracts Top Talent
    \item \textbf{23}: Develops Talent
    \item \textbf{24}: Values Differences
    \item \textbf{26}: Communicates Effectively
    \item \textbf{29}: Persuades
\end{itemize}
\end{APPENDICES}

\end{document}